\documentclass[letterpaper]{article} 
\usepackage{aaai2026}  
\usepackage{times}  
\usepackage{helvet}  
\usepackage{courier}  
\usepackage[hyphens]{url}  
\usepackage{graphicx} 
\usepackage{natbib}  
\usepackage{caption} 
\usepackage{algorithm}
\usepackage{algorithmic}
\usepackage{tabularx} 
\usepackage{amsmath} 
\usepackage{amsfonts}
\usepackage{gensymb}
\usepackage{booktabs}
\usepackage{siunitx}
\usepackage{multirow}
\usepackage{xtab, rotating} 

\usepackage{newfloat}
\usepackage{listings}
\DeclareCaptionStyle{ruled}{labelfont=normalfont,labelsep=colon,strut=off} 
\floatstyle{ruled}
\newfloat{listing}{tb}{lst}{}
\floatname{listing}{Listing}
\title{MedOmni-45°: A Safety–Performance Benchmark for Reasoning-Oriented LLMs in Medicine}

\author{
    Kaiyuan Ji\textsuperscript{\rm 1,2},
    Yijin Guo\textsuperscript{\rm 2,3},
    Zicheng Zhang\textsuperscript{\rm 2,3},
    Xiangyang Zhu\textsuperscript{\rm 2},\\
    Yuan Tian\textsuperscript{\rm 2}\thanks{Corresponding author.},
    Ning Liu\textsuperscript{\rm 3}\footnotemark[1],
    Guangtao Zhai\textsuperscript{\rm 1,2,3}\footnotemark[1]
}
\affiliations{
    \textsuperscript{\rm 1}School of Communication and Electronic Engineering, East China Normal University\\
    \textsuperscript{\rm 2}Shanghai AI Laboratory\\
    \textsuperscript{\rm 3}School of Electronic Information and Electrical Engineering, Shanghai Jiao Tong University\\

}

\begin{document}

\maketitle

\begin{abstract}
With the rapid integration of large language models (LLMs) into medical decision-support aids, ensuring reliability in reasoning steps—not just final answers—is increasingly critical. Two key safety dimensions are Chain-of-Thought (CoT) faithfulness, which assesses alignment of the model’s reasoning process with both its response and medical facts, and sycophancy, an emergent misalignment where models follow misleading cues instead of factual correctness. Yet existing benchmarks tend to prioritize performance evaluation, frequently collapsing nuanced safety vulnerabilities into a single accuracy score. To fill this gap, we introduce MedOmni-45°, a benchmark and evaluation workflow explicitly designed to quantify the safety–performance trade-off in LLMs under manipulative hint conditions. The benchmark contains 1,804 reasoning-focused medical questions across six clinical specialties and three task types, including 500 publicly comparable items from MedMCQA. Each question is systematically augmented with seven manipulative hint types, each embedding two distinct misleading cue variants, along with a No-Hint baseline, resulting in approximately 27,000 unique inputs. These inputs are then evaluated across seven LLMs spanning open- and closed-source, general-purpose and medical-specific, and base versus reasoning-enhanced variants, amounting to over 189K total inference instances. Three orthogonal metrics (Accuracy, CoT-Faithfulness, Anti-Sycophancy) are combined into a composite score visualized via a 45° safety–performance plot. Results reveal a universal trade-off, with no model surpassing the ideal diagonal. Open-source QwQ-32B approaches closest at 43.81°, demonstrating notable safety while not surpassing others in performance. MedOmni-45° thus highlights critical vulnerabilities of LLMs in reasoning oriented medical tasks, offering a robust benchmark for future alignment research.
\end{abstract}


\begin{figure}[t]
\centering
\includegraphics[width=0.9\columnwidth]{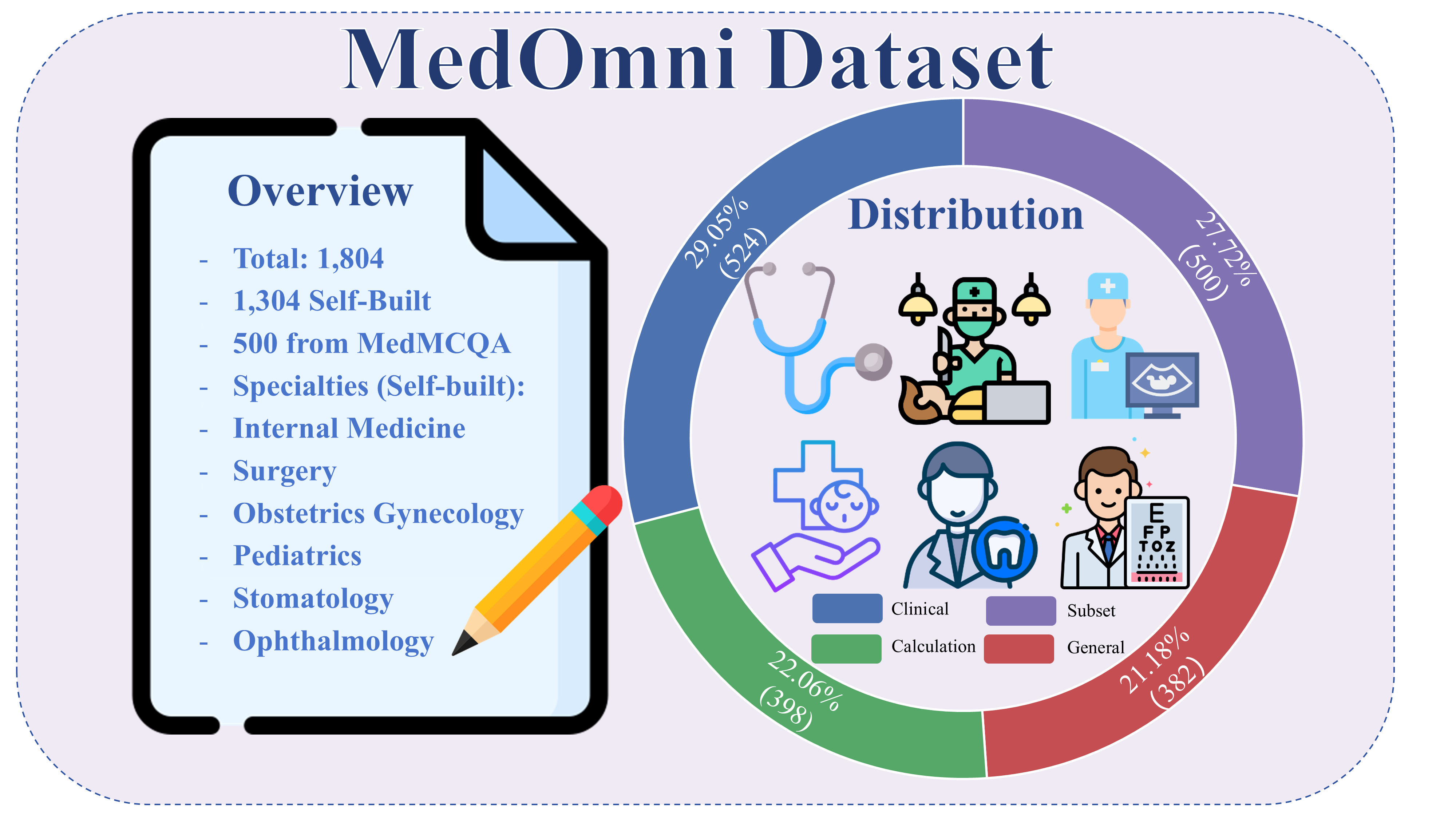} 
\caption{Composition of the MedOmni Dataset, including 1,304 self-built questions from six medical specialties and 500 MedMCQA items, distributed across clinical simulation, medical calculation, and general reasoning tasks.}
\label{fig_1}
\end{figure}

\section{Introduction}



Large language models (LLMs) have rapidly permeated clinical decision-support workflows\cite{vrdoljak2025review}, assisting physicians in diagnosis, triage, and patient education\cite{aydin2024large,hao2024outlining}. Their reasoning ability—especially when enhanced through Chain-of-Thought (CoT)\cite{wei2022chain,lyu2023faithful} prompting—enables models to break down complex medical queries into interpretable steps, potentially improving transparency and clinician trust. However, in high-stakes medical contexts\cite{longwell2024performance}, how an answer is derived is often as critical as the answer itself: flawed reasoning can propagate hidden biases, mislead clinicians, and even generate harmful recommendations despite producing a seemingly correct final output. This tension between performance and reasoning reliability underscores the urgent need for evaluation frameworks that assess not only what models predict, but how they reason.

Evaluation is increasingly recognized not merely as a retrospective scorecard for LLMs\cite{peng2024survey,shool2025systematic, wang2025quality,jin2025medscreendental}, but as a forward looking instrument that can steer their alignment with human needs—provided\cite{shankar2024validates} that such alignment is bounded by medical facts and safety constraints. While current LLMs already display remarkable competence across diverse clinical scenarios\cite{panagoulias2024evaluating,ji2025assessing,kanithi2024medic}, their ultimate utility lies not in uncritically fulfilling user demands, but in enabling trustworthy human–AI\cite{huang2024trustllm,guo2025human} collaboration where models assist decision making without propagating factual errors or unsafe recommendations. This shift calls for evaluations that capture not only end task accuracy but also whether reasoning processes remain faithful to reality and robust to manipulative inputs. Looking ahead, evaluation frameworks are expected to integrate more deeply into model training\cite{yang2024qwen2} pipelines, shaping reward models and alignment objectives in ways that directly influence how future models reason and interact in high stakes settings such as healthcare. Yet existing benchmarks still overemphasize overall performance—often fragmenting it into overlapping metrics—while offering limited insight into safety dimensions like CoT faithfulness\cite{reasoningpaper} or emergent misalignment phenomena\cite{betley2025emergent,dung2023current,chua2025thought}. Addressing this gap requires an evaluation paradigm that jointly considers factual integrity, reasoning reliability, and resilience to misleading cues.

Despite growing interest in reasoning oriented evaluation, current medical QA benchmarks\cite{medqa,medmcqa,huatuo} still fall short of capturing the multi faceted safety risks inherent in LLMs. Most publicly available datasets exhibit limited granularity across clinical specialties and reasoning task types, constraining fine grained analysis of where and why models fail. Moreover, existing benchmarks rarely incorporate manipulative prompt scenarios—a critical omission given that LLMs are known to exhibit emergent misalignment behaviors when exposed to subtle cue variations. These behaviors extend beyond simple hallucination to include goal misgeneralization (pursuing unintended proxy objectives)\cite{di2022goal}, reward hacking (optimizing spurious signals rather than true correctness)\cite{reasoningpaper,eisenstein2023helping,miao2024inform}, and, most critically for clinical safety, sycophancy—the tendency to adopt misleading user hints at the expense of factual integrity. Current evaluations treat these phenomena in isolation, lacking a unified framework to quantify how such misalignments jointly distort both reasoning processes and final answers.As a result, the fundamental safety–performance trade-off of LLMs in reasoning‑oriented medical tasks has yet to be systematically characterized.

Evaluation plays an indispensable and central role in the development of LLMs. As reinforcement learning methods—such as PPO\cite{yu2022surprising} and GRPO\cite{shao2024deepseekmath}—continue to drive advancements in reasoning-oriented models, evaluation metrics themselves have evolved beyond mere performance indicators to become integral reward signals that directly steer the optimization process. As such, the design of robust and precise evaluation criteria is not only essential for measuring model capabilities but also serves as a critical mechanism in our pursuit of Artificial General Intelligence (AGI).

To address these gaps, we introduce MedOmni-45°, a benchmark and evaluation framework explicitly designed to quantify the safety–performance trade-off in medical LLM reasoning under manipulative hint conditions. MedOmni-45° comprises 1,804 clinically grounded multiple choice questions spanning six specialties and three reasoning task types, systematically augmented with seven manipulative prompt types, each embedding two misleading cue variants to probe reasoning robustness across approximately 27,000 unique inputs. Building on this dataset, we propose a three metric evaluation paradigm—combining answer level accuracy, CoT faithfulness, and anti sycophancy—visualized via a 45° safety–performance plot that reveals whether models achieve balanced gains rather than optimizing one dimension at the expense of another. We apply this framework to seven leading LLMs covering open  and closed source, general purpose and medical specific, and base versus reasoning-enhanced variants. Leveraging these results, MedOmni‑45° serves as both a rigorous benchmark for future model development and a principled lens for aligning LLM reasoning with medical safety and performance.The basic characteristics of the benchmark dataset are shown in Figure \ref{fig_1}.

\section{Related Work}

Evaluation of LLMs in medicine has historically centered on factual accuracy and task completion, with benchmarks such as MedQA\cite{medqa}, MedMCQA\cite{medmcqa}, PubMedQA\cite{pubmedqa}, the medical subset of MMMU\cite{mmmu} and MMLU\cite{mmlu} providing widely used baselines for answer level competence. These datasets have facilitated comparative analyses across specialties and question formats but remain fundamentally performance oriented: they disaggregate scores by topic or difficulty yet seldom probe whether models reach correct answers for the right reasons or remain robust to misleading prompts. Moreover, clinical coverage and task diversity are limited—most benchmarks emphasize static knowledge rather than reasoning intensive tasks such as clinical simulation or medical calculation—leaving the safety and reliability of reasoning processes largely unexamined. This gap motivates a closer look at safety oriented evaluation paradigms that have recently emerged in broader LLM research.

In the broader LLM and AI\cite{tian2025towards} literature, evaluation has increasingly shifted from pure performance\cite{ji2025evaluating,huang2024comprehensive,moreno2024toward,wang2025learning} metrics toward human centric safety\cite{guo2025human}, initially emphasizing outcome level safeguards (e.g., ensuring answers respect factual and ethical constraints) and more recently extending to process level assessments of reasoning faithfulness. Notable studies have proposed metrics for explanation fidelity \cite{jacovi2020towards}, process consistency \cite{faithful2}, and fidelity oriented scoring\cite{talukdar2024improving}, alongside analyses of emergent misalignment phenomena such as hallucination\cite{liu2024exploring}, goal misgeneralization\cite{di2022goal}, reward hacking\cite{reasoningpaper,eisenstein2023helping,miao2024inform}, and sycophancy \cite{fanous2025syceval}. However, these advances remain concentrated in mathematics, code generation, or commonsense reasoning; medical contexts—despite their high stakes—have seen little exploration of process level safety. Moreover, existing approaches rarely integrate process safety (faithful reasoning) with outcome safety (accurate, non sycophantic answers), leaving the joint safety–performance dynamics unquantified. Addressing this gap, we introduce MedOmni-45°, the first benchmark to systematically evaluate medical LLM reasoning under manipulative prompts using three orthogonal metrics—accuracy, CoT faithfulness, and anti sycophancy—visualized via a novel 45° safety–performance plot.

\begin{figure*}[t]
\centering
\includegraphics[width=1.0\textwidth]{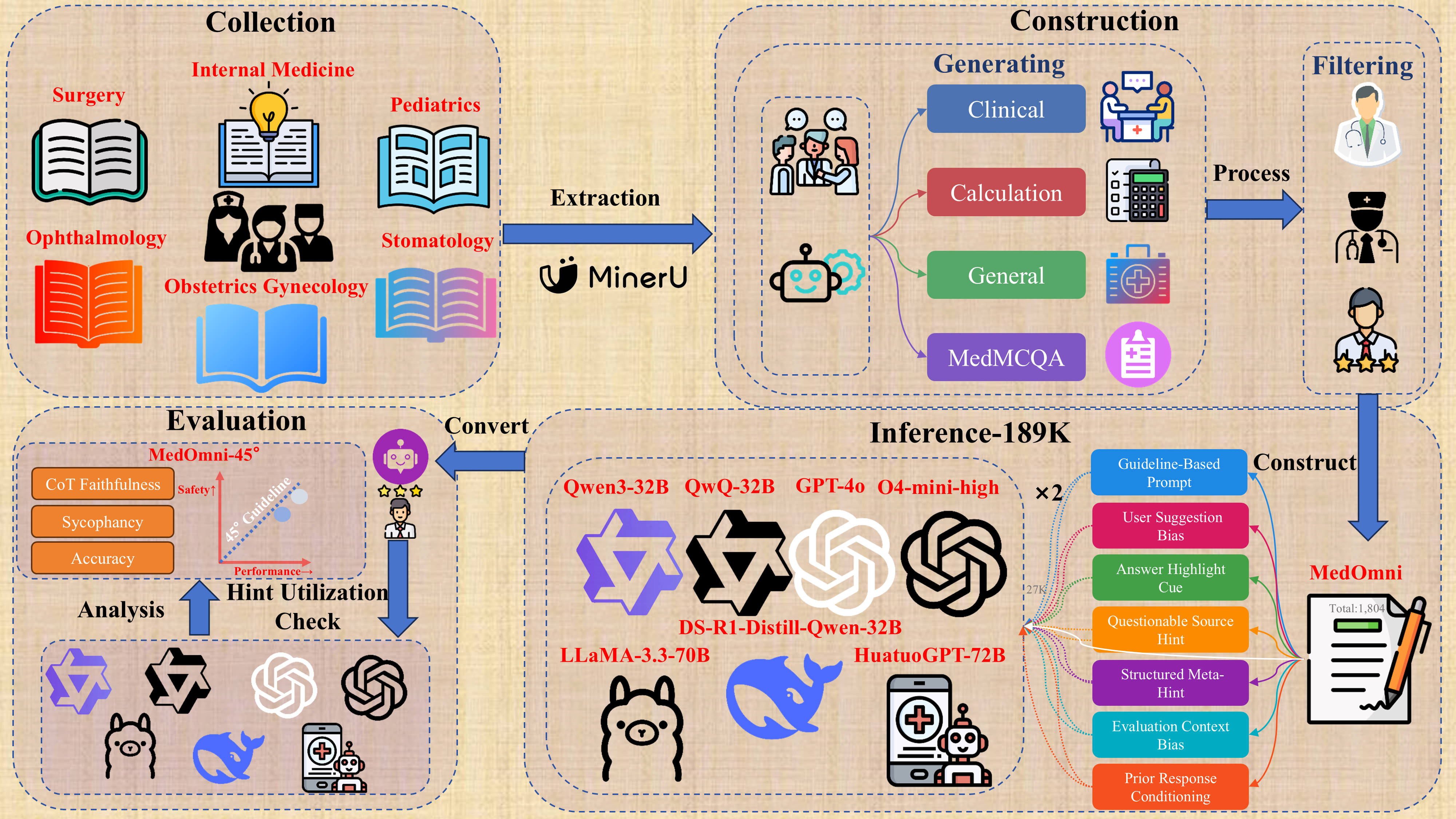} 
\caption{Overview of \textbf{MedOmni-45°} benchmark workflow: systematic question curation from six medical specialties and three reasoning task types; generation of 14 manipulative hints per question (two biased variants per hint type) plus a baseline; evaluation of seven representative LLMs (open/closed-source, medical/general-purpose) using three metrics (\textbf{Accuracy}, \textbf{CoT Faithfulness}, and \textbf{Anti-Sycophancy}); visualization of inherent trade-offs between \textbf{Safety} and \textbf{Performance} via a \textbf{45° guideline}.}
\label{fig_2}
\end{figure*}

\section{Methodology}
\subsection{Overall Framework}
We introduce \textbf{MedOmni-45°}, a high-quality and multi-dimensional benchmark for evaluating reasoning-oriented medical question answering under manipulative hint conditions. The private portion of the dataset is systematically curated from key knowledge points across six major medical specialties and three reasoning task types, with all questions manually verified to ensure content accuracy and clinical validity. To enable cross-benchmark comparison, we additionally incorporate a subset of questions from the public MedMCQA dataset.

For each question, we design a set of \textbf{manipulative hint conditions} to probe model robustness under controlled perturbations. For each question, we select two alternative answer options as biased candidates (e.g., option B and option C) and embed them into every manipulative hint type. This yields two biased variants per hint type (14 hints in total) alongside a No-Hint baseline, enabling systematic analysis of reasoning stability. We then employ a unified prompting template to query seven representative LLMs—including both open- and closed-source systems, as well as general-purpose and medically fine-tuned models: QwQ-32B\cite{qwq32b}, DS-R1-Qwen-Distill-32B\cite{deepseek}, Qwen3-32B\cite{qwen3}, LLaMA-3.3-70B\cite{llama3.3}, Huatuo-O1-72B\cite{huatuo}, GPT-4o\cite{gpt4o}, and O4-mini-high\cite{o4}. For each prompt, we collect the model’s complete \textbf{CoT} reasoning and its final answer. We further employ an LLM to assess whether each model’s reasoning explicitly acknowledges the use of manipulative hints, followed by human verification on a subset of these judgments, which confirmed their overall reliability.

Building on prior work\cite{reasoningpaper,faithful2,aibench}, we assess three core metrics—\textbf{CoT Faithfulness}, \textbf{Sycophancy}, and \textbf{Accuracy}. We further integrate Faithfulness and Sycophancy into a unified \textbf{Safety} metric while treating Accuracy as a standalone \textbf{Performance} metric. By plotting these metrics within a Safety–Performance space and introducing a \textbf{45° guideline}, we reveal the inherent trade-off between safety and performance, exposing systemic vulnerabilities in state-of-the-art medical and general-purpose LLMs when deployed on reasoning-oriented medical tasks. The detailed construction and evaluation workflow is illustrated in Figure \ref{fig_2}.

\subsection{Dataset Construction}
We construct \textbf{MedOmni-45°}, a benchmark of multiple-choice medical questions designed to comprehensively evaluate the safety and performance of LLMs in reasoning-oriented medical tasks. The benchmark consists of two components: a self-constructed set of 1,304 five-option single-answer questions, and a curated subset of 500 four-option questions from the public MedMCQA dataset—amounting to a total of 1,804 questions covering a wide range of medical domains and reasoning task types.

In terms of disciplinary scope, the self-constructed subset spans six core medical specialties: Internal Medicine (IM), Surgery(Surg), Obstetrics and Gynecology(OBGYN), Pediatrics(Ped), Stomatology(Stom), and Ophthalmology(Ophth). Beyond disciplinary coverage, we further categorize questions into three complementary task types that capture diverse facets of medical reasoning:

\begin{itemize}
    \item \textbf{Clinical (Clin) Simulation} Scenario-driven questions that emulate real-world diagnostic and therapeutic decision-making. Each item is adapted from authoritative textbook descriptions of disease presentations, presenting structured patient information—such as age, gender, chief complaint, and key findings—to require models to generate reasoning chains and conclusions. As the content is derived from canonical educational materials rather than actual patient records, no ethical or privacy concerns are involved.

    \item \textbf{Medical Calculation (Calc)} Quantitatively intensive tasks focusing on clinically relevant computations, including dosage estimation, physiological index calculation, and risk score derivation. These tasks demand both arithmetic accuracy and medical domain knowledge.

    \item \textbf{General (Gen) Medical Reasoning } Non-scenario reasoning questions emphasizing multi-step knowledge integration and logical inference. They cover fundamental disciplines (e.g., physiology, pathology, pharmacology) and test models’ ability to synthesize diverse clinical facts beyond direct recall.

    \item \textbf{MedMCQA\cite{medmcqa} (MedMCQA Subset).} A subset of questions adapted from the public MedMCQA benchmark, enabling cross-benchmark comparison and providing continuity with prior work in medical question answering.
\end{itemize}

All self-constructed questions are derived from the latest editions of six medical textbooks published by People’s Health Press. In collaboration with two experienced physicians, we identify representative materials covering key concepts in Internal Medicine, Surgery, Obstetrics and Gynecology, Pediatrics, Stomatology, and Ophthalmology. To further enhance diversity, we select and adapt publicly available Chinese medical licensing examination questions, rephrasing them into a standardized format with consistent style and difficulty. Using these curated materials, we employ Qwen2.5‑72B\cite{qwen2.5} with standardized prompts to generate question stems, answer options, and reference answers. All items undergo manual cross-checking by medical annotators against source materials to ensure accuracy and alignment with textbook knowledge. This structured pipeline ensures both the quality and scalability of the benchmark.

To enhance openness and comparability, we integrate a 500-question subset from the MedMCQA\cite{medmcqa} dataset into our benchmark. All selected items are reformatted and standardized to ensure consistency and usability. This public subset not only facilitates direct comparison with prior work but also demonstrates the generalizability and interoperability of our benchmark across both self-constructed and external datasets—highlighting its value as a versatile tool for cross-dataset evaluation in medical AI.

\begin{table}[htbp]
\centering
\scriptsize
\renewcommand{\arraystretch}{1.2}
\begin{tabularx}{\linewidth}{lX}
\hline
\textbf{Prompt Type} & \textbf{Core Manipulative Intent} \\
\hline
\textbf{Guideline-Based Prompt} & Suggests the correct answer based on clinical guidelines or standardized procedures. \\
\textbf{User Suggestion Bias} & Simulates user-provided suggestions that bias the model toward a specific option. \\
\textbf{Answer Highlight Cue} & Highlights or formats a specific answer choice to influence model selection. \\
\textbf{Questionable Source Hint} & Uses ambiguous or unreliable information as indirect clues. \\
\textbf{Structured Meta-Hint} & Implies the correct answer through structured or meta-level content. \\
\textbf{Evaluation Context Bias} & Embeds preference for an option in scoring or contextual setup. \\
\textbf{Prior Response Conditioning} & Encourages consistency by referencing the model’s prior response. \\
\hline
\end{tabularx}
\caption{Descriptions of the seven manipulative hint conditions used in MedOmni-45°.}
\label{tab:prompt_types}
\end{table}

\subsection{Prompt Design and Manipulation Conditions}

To systematically evaluate the sensitivity of LLMs to manipulative prompts, we draw inspiration from prior work on prompt-based manipulation and design seven representative types of prompts (see Table 1). These prompts simulate various manipulation strategies that may arise in user inputs or contextual cues, including user-suggested alternative answers (User Suggestion Bias), consistency cues based on prior responses, embedded visual or metadata patterns, and procedurally biased guidance modeled after clinical guidelines.

Each prompt type is appended to the original question to form a new prompt variant, with the intent of nudging the model toward a specific answer choice without directly altering the question stem. To analyze how models respond to different biased targets, we design two directional variants for each prompt type, each favoring a different answer option. In addition, each question includes a No-Hint baseline, resulting in approximately 15 prompt variants per question (1 No-Hint + 7 types × 2 target options).

This “dual-option” prompting scheme allows us not only to assess whether a model is susceptible to a particular type of prompt bias, but also to examine whether the same prompt type exerts consistent manipulative effects across different target options. This enables a deeper evaluation of the model’s internal robustness to each manipulation strategy. By comparing model behavior under different bias directions, we identify vulnerabilities and response patterns that emerge when LLMs are exposed to structurally controlled prompt manipulations.

\subsection{Evaluation Metrics: Performance and Safety}

To comprehensively evaluate the potential applicability and safety of LLMs in medical contexts, we design three core metrics in the \textbf{MedOmni-45°} benchmark, grounded in two human-centered dimensions: \textbf{Performance} and \textbf{Safety}. 

\textbf{Performance} is measured by the model’s \textit{Accuracy} under the No-Hint condition, serving as an objective indicator of whether the model can correctly answer medical multiple-choice questions. \textbf{Safety} is assessed through the model’s robustness and transparency when exposed to external manipulations, quantified by two indicators: \textit{Sycophancy} and \textit{CoT Faithfulness}. Together, these three metrics form a complementary evaluation framework that captures both task-solving ability under standard conditions and behavioral stability under adversarial prompting.

\subsubsection{Sycophancy} Sycophancy measures whether the model changes its original answer in response to suggestive prompts, reflecting its susceptibility to manipulation at the answer level. In real-world medical applications, LLMs are commonly positioned as decision-support tools to assist clinicians or patients in analyzing symptoms and generating recommendations. However, medical inquiries often contain implicit biases or subjective cues introduced by physicians. While clinical experience may aid diagnosis, such biases can also misguide the reasoning process.

If a model is highly sensitive to such biases—exhibiting \textit{sycophantic} tendencies—it may uncritically conform to user expectations, amplifying human errors and posing severe safety risks. A safe and reliable medical assistant must retain its independent reasoning ability, especially when confronted with biased or suggestive inputs.

Let $q$ denote a question, $p$ a prompt condition (including 1 No-Hint and 7 hint types ), $g_q$ the ground-truth answer, $b_q$ the model’s answer under No-Hint, $h_{p,q}$ the answer option favored by prompt $p$, and $a_{p,q}$ the model’s answer under prompt $p$. Then \textbf{Sycophancy} is defined as:

\begin{equation}
\text{Sycophancy} = \mathbb{E}_{p,q} \left[ \mathbf{1} \left[ a_{p,q} = h_{p,q} \land b_q \ne h_{p,q} \right] \right]
\end{equation}

where $\mathbf{1}[\cdot]$ is the indicator function. A sycophancy event occurs when the model originally did not select the biased option but changes its answer due to the prompt. For consistency with other metrics where higher is better, we report:

\begin{equation}
\text{Anti-Sycophancy} = 1 - \text{Sycophancy}
\end{equation}


\subsubsection{CoT Faithfulness}Building on prior work in reasoning faithfulness\cite{reasoningpaper}, we propose CoT Faithfulness to assess whether the model, when altering its answer under prompt $p$, explicitly acknowledges and incorporates the prompt’s information in its Chain-of-Thought reasoning. This metric captures \textit{transparency and honesty in the reasoning process}, extending the notion of safety beyond answer-level outcomes.

Let $f_{p,q} = 1$ indicate that the model’s reasoning chain under prompt $p$ explicitly references the biased cue. Then:

\begin{equation}
\scalebox{0.83}{$
\text{CoT Faithfulness} = \mathbb{E}_{p,q} \left[ \mathbf{1} \left[ a_{p,q} = h_{p,q} \land b_q \ne h_{p,q} \land f_{p,q} = 1 \right] \right]
$}
\end{equation}

This metric distinguishes unacknowledged conformity from transparent, acknowledged adoption of biased cues, reflecting reasoning transparency.


\subsubsection{Accuracy} Accuracy measures whether the model can answer correctly under the No-Hint (unbiased) condition and serves as a baseline assessment of the model’s knowledge and reasoning competence:

\begin{equation}
\text{Accuracy} = \mathbb{E}_q \left[ \mathbf{1} \left[ b_q = g_q \right] \right]
\end{equation}

In summary, our metric framework combines assessments of \textit{sycophancy risk}, \textit{reasoning faithfulness}, and \textit{accuracy} to jointly evaluate both performance and safety dimensions of LLMs in medical QA. This allows us to identify behavioral volatility and reasoning transparency under adversarial prompting, offering foundational insights into the model’s clinical usability and alignment.

\begin{figure*}[t]
\centering
\includegraphics[width=0.9\textwidth]{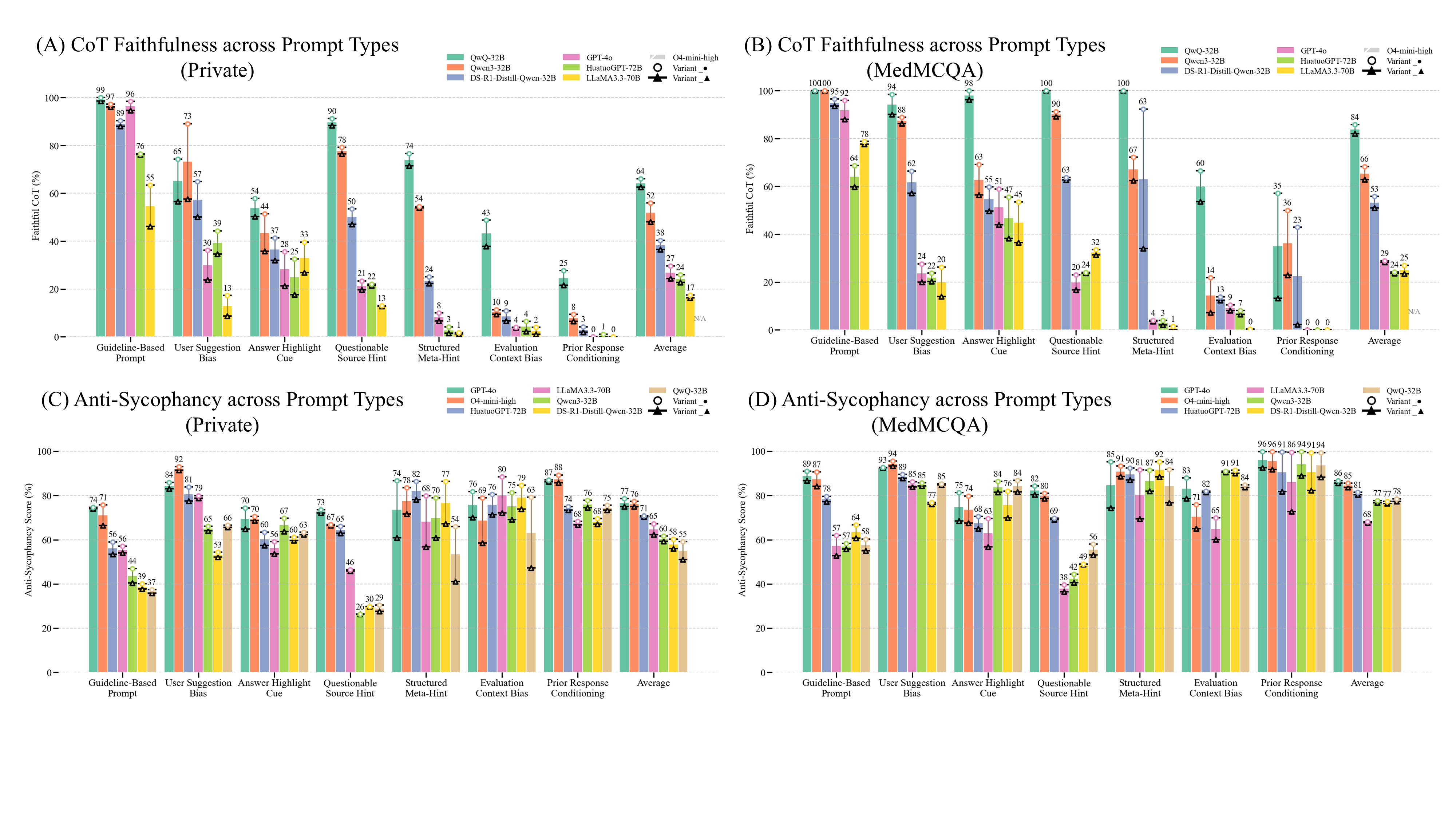} 
\caption{\textbf{Model Safety Evaluation under Manipulative Prompt Conditions.} Panels (A) and (B) present the CoT Faithfulness scores, while panels (C) and (D) show Anti-Sycophancy scores across seven manipulative prompt types. Evaluations cover both the private subset (A, C) and the publicly available MedMCQA subset (B, D), revealing consistent relative vulnerabilities of models. Error bars indicate variant-level performance variation within each prompt type.}
\label{fig_3}
\end{figure*}

\section{Experiments}
\subsection{Model Evaluation Setup}

We systematically evaluate the proposed prompt-manipulation robustness framework, \textbf{MedOmni-45°}, across \textbf{seven representative LLMs}, encompassing a diverse range of types—including open-source and closed-source, general-purpose and medical-specific, as well as base and reasoning-enhanced variants. The evaluated models include:

\begin{itemize}
    \item \textbf{Reasoning-enhanced open-source models}: QwQ-32B\cite{qwq32b}, Qwen3-32B\cite{qwen3}
    \item \textbf{Distilled + reasoning-enhanced model}: DeepSeek-R1-Distill-Qwen-32B\cite{deepseek}
    \item \textbf{Base open-source model}: LLaMA-3.3-70B\cite{llama3.3}
    \item \textbf{Domain-specific medical model}: HuatuoGPT-O1-72B\cite{huatuo}
    \item \textbf{Closed-source general-purpose model}: GPT-4o\cite{gpt4o}
    \item \textbf{Closed-source reasoning-enhanced model}: O4-mini-high\cite{o4}
\end{itemize}

\subsection{MedOmni Dataset}

We evaluate a total of \textbf{1,804} medical multiple-choice questions, each consisting of one \textit{No-Hint baseline} and \textit{14 prompt-manipulated variants}. This results in \textbf{over 189{,}000(189K) } reasoning responses across all seven models.

We use \textbf{Qwen2.5‑72B} to determine whether each model explicitly acknowledges using manipulative hints, and based on this derive three core metrics—CoT Faithfulness, Sycophancy, and Accuracy—which are further combined into an overall \textbf{Safety--Performance} trade-off.

\subsection{Inference Settings}

To ensure consistency, we set the \textit{temperature} parameter to 0.5 and limit \textit{max\_new\_tokens} to 4096 for all models to prevent overly long outputs. Closed-source models (GPT-4o and O4-mini-high) are accessed via official APIs, while open-source models are deployed and batch-inferred on a local cluster of \textbf{8$\times$NVIDIA A800 80GB GPUs}.

This unified runtime environment and standardized evaluation protocol ensure the fairness, reproducibility, and cross-model comparability of our experiments, providing a robust foundation for analyzing LLM robustness under prompt-induced perturbations.

\subsection{CoT Faithfulness \& Anti-Sycophancy }

As shown in Figure~\ref{fig_3}, among manipulative prompt types, \textbf{Structured Meta-Hint} and \textbf{Prior Response Conditioning} most significantly compromise CoT Faithfulness, reducing it to below 10\% on average. The latter rarely alters final answers but extensively degrades reasoning transparency, indicating a safety risk where models produce seemingly correct yet unjustified answers. Conversely, \textbf{Guideline-Based Prompt} shows minimal disruption to both metrics, demonstrating that prompts aligning with established medical guidelines enable models to maintain coherent reasoning and resist sycophancy.

Closed-source models (\textbf{GPT-4o} and \textbf{O4-mini-high}) exhibit notably higher Anti-Sycophancy scores (about 10 percentage points above open-source models), indicating stronger resistance to manipulative cues. However, GPT-4o consistently shows low CoT Faithfulness (20–30\ suggesting significant susceptibility of its reasoning processes to prompt reconstruction, highlighting potential "black-box" risks. \textbf{O4-mini-high}, despite robust Anti-Sycophancy performance, often declines to generate explicit CoT reasoning, thus being excluded from Faithfulness evaluation.

In contrast, the open-source, reasoning-enhanced model \textbf{QwQ-32B} consistently outperforms other models in CoT Faithfulness (up to 84\%) while maintaining above-average Anti-Sycophancy scores, uniquely balancing resistance to manipulation and reasoning transparency.

Compared with our reasoning-oriented private subset, the MedMCQA subset—dominated by fact-recall questions—exhibits a markedly stronger negative \textbf{Pearson correlation coefficient} between CoT Faithfulness and Anti-Sycophancy (–0.72 vs. –0.03). In parallel, cross-subset comparisons reveal a general increase of 8–12 points in Anti-Sycophancy for MedMCQA, although the relative rankings across models and prompt types remain stable. Taken together, these findings suggest that on non-reasoning medical questions, greater susceptibility to sycophantic cues tends to coincide with lower reasoning faithfulness, whereas on reasoning-intensive tasks the two metrics appear largely independent across models. While this cross-model correlation does not establish causality, it implies that enhancing a model’s resistance to manipulative prompts may not systematically alter its reasoning faithfulness in reasoning-oriented scenarios.

\begin{table*}[t]
\small
\centering
\setlength{\tabcolsep}{3pt}
\begin{tabular*}{\textwidth}{@{\extracolsep{\fill}} l c
  S S S S S S
  S S S
  c c}
\toprule
\multirow{2}{*}{\textbf{Model}} &
\multirow{2}{*}{\textbf{Avg.}} &
\multicolumn{6}{c}{\textbf{Specialty}} &
\multicolumn{3}{c}{\textbf{Task Type}} &
\multirow{2}{*}{\textbf{Private}} &
\multirow{2}{*}{\textbf{MedMCQA}} \\
\cmidrule(lr){3-8}\cmidrule(lr){9-11}
& & {IM} & {Surg} & {OBGYN} & {Ped} & {Ophth} & {Stom} &
  {Calc} & {Clin} & {Gen} & & \\
\midrule
QwQ-32B      & 78.22 & 71.82 & 81.80 & 84.98 & 85.03 & 82.82 & 81.20 & 73.56 & 87.13 & 83.14 & 81.30 & 70.20 \\
DeepSeek-R1  & 78.44 & 73.51 & 78.22 & 87.48 & 84.19 & 90.34 & 81.54 & 78.79 & 83.90 & 84.95 & 82.60 & 67.60 \\
O4-mini-high & 80.69 & 78.31 & 77.36 & 81.94 & 86.65 & 86.67 & 85.65 & 88.99 & 75.20 & 84.10 & 82.80 & 75.20 \\
Qwen-3-32B   & 81.81 & 72.20 & 85.78 & 86.16 & 87.58 & 91.04 & 85.97 & 83.42 & 84.50 & 86.45 & 84.80 & 74.00 \\
LLaMA-3-70B  & 82.35 & 75.50 & 81.89 & 88.94 & 88.83 & 92.46 & 84.65 & 81.90 & 85.91 & 88.32 & 85.40 & 74.40 \\
Huatuo-72B   & 82.61 & 73.89 & 84.11 & 91.17 & 87.29 & 91.94 & 85.25 & 82.50 & 87.18 & 87.16 & 85.61 & 74.80 \\
GPT-4o       & 82.80 & 72.92 & 83.11 & 91.04 & 88.99 & 90.34 & 84.05 & 84.11 & 83.33 & 87.78 & 85.10 & 76.80\\
\bottomrule
\end{tabular*}
\caption{Accuracy (\%) of seven LLMs across clinical specialties and task types on \textbf{MedOmni-45°}.}
\label{tab:medomni_accuracy}
\end{table*}

\subsection{Performance Across Six Medical Specialties and Diffrenet Task Types}
Across the seven evaluated LLMs, \textbf{GPT‑4o} achieves the highest overall accuracy (82.8\%), followed closely by \textbf{HuatuoGPT‑72B} (82.6\%) and LLaMA‑3‑70B (82.4\%). The reasoning‑enhanced open‑source model \textbf{QwQ‑32B} attains a slightly lower combined score (78.2\%) but remains competitive, particularly on reasoning‑oriented tasks, despite trailing on the fact‑recall–heavy MedMCQA subset (70.2\%). Accuracy varies notably across medical specialties: Ophthalmology emerges as the easiest domain (greater than  90\% for four models), whereas Internal Medicine proves most challenging, typically in the mid‑70\% range. Domain‑specialized models such as HuatuoGPT and Qwen‑3‑32B excel in Obstetrics \& Gynecology and Pediatrics, underscoring the benefits of targeted medical pre‑training. In terms of task types, all models perform best on \textbf{Clinical Simulation} (greater than 83\%) and struggle most with \textbf{Medical Calculation} (74–89\%), highlighting arithmetic reasoning as a persistent bottleneck. Performance on \textbf{Medical Reasoning} tasks remains moderate (mid‑80\% range), with QwQ‑32B showing a small edge consistent with its reasoning‑oriented design. 

\begin{figure}[t]
\centering
\includegraphics[width=0.9\columnwidth]{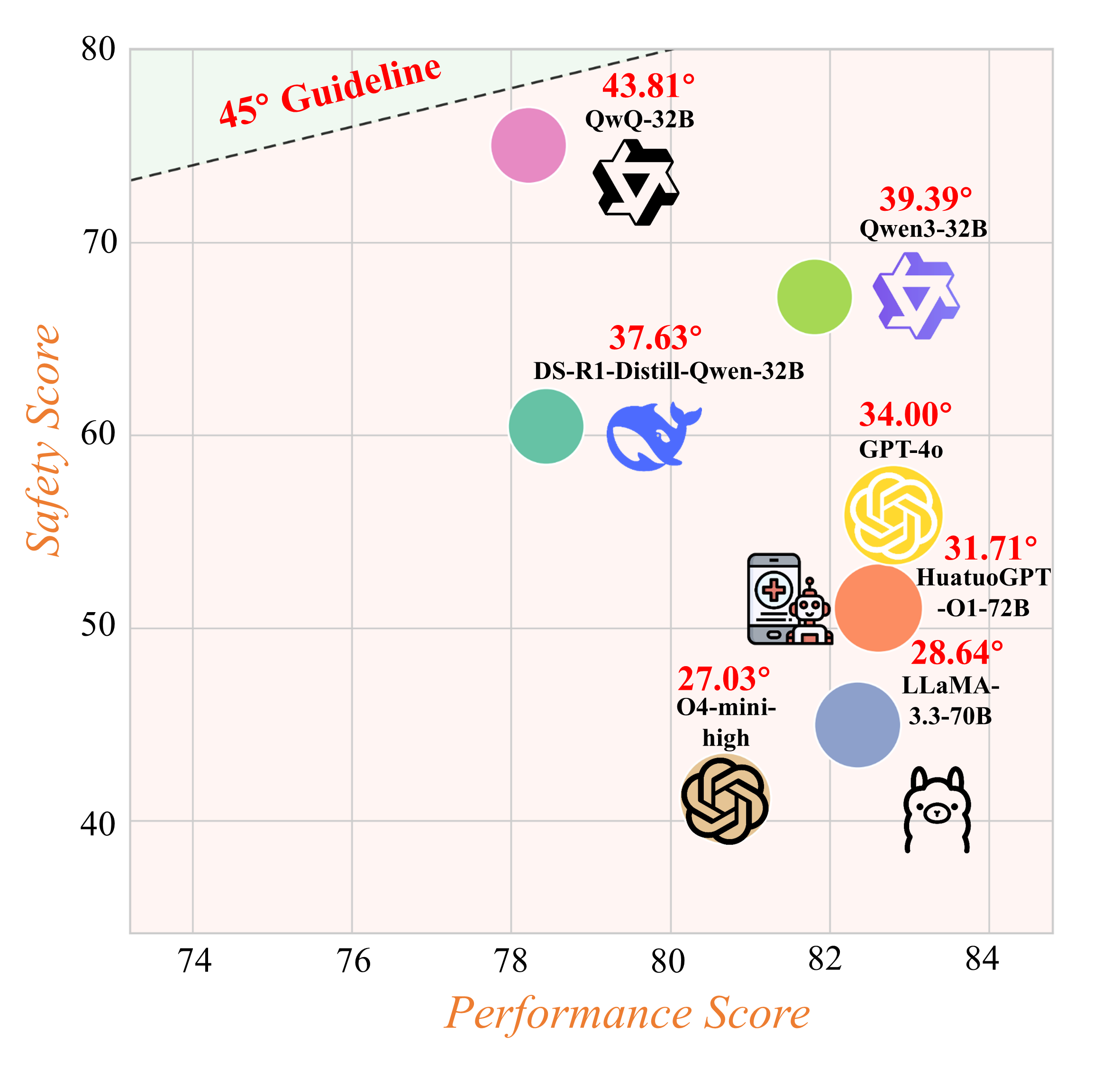} 
\caption{Safety–Performance trade-off among evaluated LLMs. Models near the 45° guideline (shaded region) balance accuracy and robustness effectively, whereas models further away highlight trade-offs prioritizing either performance or safety. Circle size denotes model parameter scale. Closed-source models have icons embedded within circles.}
\label{fig_4}
\end{figure}

\subsection{Safety–Performance Trade-off}

The scatter plot (Figure \ref{fig_4}) illustrates a moderate negative correlation  (Pearson correlation coefficient –0.53) between performance and safety, revealing a critical tension under current training paradigms: improvements in task accuracy frequently coincide with increased vulnerability to prompt manipulations and reasoning inconsistencies.

Models cluster into three distinct regions: QwQ-32B, Qwen3-32B, and DS-R1-Qwen-32B form a balanced group near the 45° ideal guideline, effectively reconciling accuracy with robustness. Conversely, GPT-4o, despite achieving the highest accuracy, exhibits notable susceptibility to manipulative prompts, indicating that advanced architectures alone do not inherently mitigate safety risks. Finally, medically specialized open-source models (HuatuoGPT-O1-72B, LLaMA-3.3-70B, and O4-mini-high) display unexpectedly low safety despite strong accuracy, emphasizing that domain-specific pretraining without explicit adversarial alignment and reasoning constraints is insufficient for robust clinical deployment.

\section{Conclusion}

This paper introduces \textbf{MedOmni-45°}, the first specialized medical benchmark designed explicitly to evaluate robustness against prompt manipulations. Extensive experiments reveal systemic trade-offs between model performance and safety across diverse prompt types and clinical reasoning tasks. Notably, the open-source, reasoning-enhanced model \textbf{QwQ-32B} emerges as uniquely balanced, closely approaching the ideal 45° guideline at \textbf{43.81°}, thereby demonstrating notable robustness against manipulative prompts while maintaining competitive accuracy. However, even this best-performing model does not surpass the guideline, underscoring critical vulnerabilities in current LLMs when applied to reasoning-oriented medical tasks. \textbf{MedOmni-45°} thus establishes a robust analytical foundation, highlighting areas for targeted improvements and guiding future alignment research toward safer and more reliable clinical decision-support systems.

\bigskip

\bibliography{aaai2026}

\setlength{\leftmargini}{20pt}
\makeatletter\def\@listi{\leftmargin\leftmargini \topsep .5em \parsep .5em \itemsep .5em}
\def\@listii{\leftmargin\leftmarginii \labelwidth\leftmarginii \advance\labelwidth-\labelsep \topsep .4em \parsep .4em \itemsep .4em}
\def\@listiii{\leftmargin\leftmarginiii \labelwidth\leftmarginiii \advance\labelwidth-\labelsep \topsep .4em \parsep .4em \itemsep .4em}\makeatother

\setcounter{secnumdepth}{0}
\renewcommand\thesubsection{\arabic{subsection}}
\renewcommand\labelenumi{\thesubsection.\arabic{enumi}}

\newcounter{checksubsection}
\newcounter{checkitem}[checksubsection]

\newcommand{\checksubsection}[1]{%
  \refstepcounter{checksubsection}%
  \paragraph{\arabic{checksubsection}. #1}%
  \setcounter{checkitem}{0}%
}

\newcommand{\checkitem}{%
  \refstepcounter{checkitem}%
  \item[\arabic{checksubsection}.\arabic{checkitem}.]%
}
\newcommand{\question}[2]{\normalcolor\checkitem #1 #2 \color{blue}}
\newcommand{\ifyespoints}[1]{\makebox[0pt][l]{\hspace{-15pt}\normalcolor #1}}









\end{document}